\documentclass[runningheads]{llncs}
\usepackage{graphicx}
\usepackage{color}
\usepackage{commath}
\begin{document}
\title{Multi-Threshold Attention U-Net (MTAU) based Model for Multimodal Brain Tumor Segmentation in MRI scans}
\titlerunning{Multi-Threshold Attention U-Net (MTAU)  for Brain Tumor Segmentation}
%
\author{Navchetan Awasthi\inst{1,2,}\thanks{The authors contributed equally to the work.}\orcidID{0000-0001-8153-2786},\\Rohit Pardasani\inst{3,*}\orcidID{0000-0002-2444-1025} \and \newline
Swati Gupta\inst{4}\orcidID{0000-0003-1541-5498}}
\authorrunning{Navchetan Awasthi et al.}
%
\institute{Massachusetts General Hospital, USA \and
Harvard University, USA\\
\email{navchetanawasthi@gmail.com}\\
 \and
General Electric Healthcare, Bangalore, India \\
\email{rohit.r.pardasani@gmail.com}\\
 \and
Indian Institute of Science, Bangalore, India \\
\email{swati375@gmail.com}}
\maketitle              
\begin{abstract}
Gliomas are one of the most frequent brain tumors and are classified into high grade and low grade gliomas. The segmentation of various regions such as tumor core, enhancing tumor etc. plays an important role in determining severity and prognosis. Here, we have developed a multi-threshold model based on attention U-Net for identification of various regions of the tumor in magnetic resonance imaging (MRI). We propose a multi-path segmentation and built three separate models for the different regions of interest. The proposed model achieved mean Dice Coefficient of 0.59, 0.72, and 0.61 for enhancing tumor, whole tumor and tumor core respectively on the training dataset.  {The same model gave mean Dice Coefficient of 0.57, 0.73, and 0.61 on the validation dataset and 0.59, 0.72, and 0.57 on the test dataset. }

\end{abstract}
\section{Introduction}

In adults, glioma is one of the frequent brain tumor originating from glial cells and infiltrating the surrounding tissue \cite{menze2014multimodal}. The gliomas are classified into two main categories depending on the severity of gliomas - 
\begin{itemize}
    \item High grade gliomas require immediate treatment and have median survival of two years or less \cite{bakas2017advancing,bakas2018identifying,ohgaki2005population,cavenee2007classification}.
    \item Low grade gliomas have life expectancy of several years, hence aggressive treatment is often delayed \cite{bakas2017advancing,bakas2018identifying,ohgaki2005population,cavenee2007classification}.
\end{itemize} 
Various neuroimaging protocols are used for evaluating the progression of the disease before and after treatment to determine the success of the treatment strategy as well as any changes in the brain. In clinical settings, these are evaluated on the basis of qualitative criteria or by quantitative measures \cite{eisenhauer2009new,wen2010updated}.

The automatic analysis of the tumor structures will produce highly accurate and reproducible measurements of the relevant structures, thus these image processing routines are of utmost importance. Since the amount of data available for processing is huge, there is a need of an automatic method for segmenting the various regions of the brain. The analysis will help in improving the treatment planning, improving the diagnosis and follow-up of individual patients for further procedures \cite{menze2014multimodal,bakas2018identifying,bakas2017segmentation,bakas2017segmentation1}. The development of the automatic procedures for segmentation is not an easy task as the lesions are defined in terms of intensity changes compared to the surrounding tissues and even the segmentation done by expert radiologist show significant variations because of the partial volume effects, bias field artifacts and intensity gradients between the adjacent structures \cite{menze2014multimodal}. The tumor structures also vary across patients in terms of extension, localization, size and thus affecting the use of strong priors for the segmentation of the anatomical structures. The modalities provide different complementary information and thus different features are involved in segmentation \cite{menze2014multimodal,bakas2017advancing}. The imaging modalities that are used to map tumor-induced tissue changes include T2 and Fluid-Attenuated Inversion Recovery (FLAIR) MRI, post-Gadolinium T1 MRI, perfusion and diffusion MRI, and Magnetic Resonance Spectroscopic Imaging (MRSI), among
others \cite{menze2014multimodal,bakas2018identifying,bakas2017advancing}.

Previously, many models have been proposed for improving the reconstruction and improving the segmentation such as U-Net, U-Net++ etc. \cite{ronneberger2015u,zhou2018unet++}. Recently, there has been focus on modified U-Net architectures such as attention-U-Net as well as hybrid U-Net architectures for image reconstruction as well as image super-resolution based enhancement \cite{awasthi2020deep,awasthi2020sinogram,ronneberger2015u,awasthi2019pa}. Similarly, many architectures have been developed using texture analysis, active contours, random forests as well as probabilistic models for tumor segmentation \cite{gordillo2013state}. 

Here, we have proposed a Multi-Threshold Attention U-Net (MTAU) \cite{oktay2018attention} based 2D model for segmentation of the multimodal brain tumor images of MRI scans into three different regions (Necrotic (NCR) and the Non-Enhancing Tumor (NET), Enhancing Tumor (ET) and Edema (ED)) by individually training three different models. The three models use identical architecture, thus we reduce the memory requirement (the 2D model being less complex than a 3D one and has lesser number of parameters) at the time of training/inference without increasing the effort of model design. Alternately, in the absence of memory constraint we can use same models in a multi-path architecture format where 2D U-Net models get trained in parallel.

\section{Methods}
\begin{figure}[!h]
\centering\includegraphics[width=\linewidth,height=18.0 em]{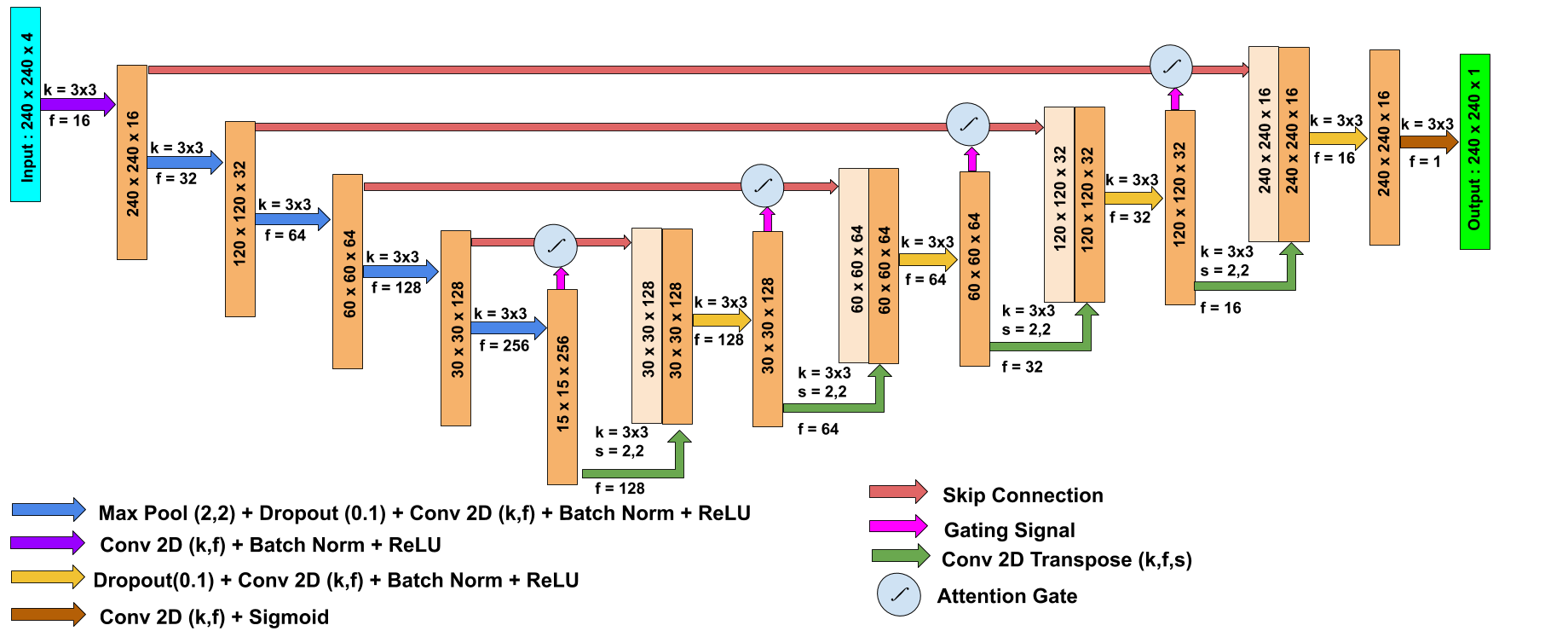}
\caption{\label{Figure-1}{Architecture of the proposed  attention U-Net model utilized in this work. The various size of filters and the corresponding connection are shown in the architecture and the binary cross entropy loss has been used for back-propagation.
}}
\end{figure}

\begin{figure}[!h]
\centering\includegraphics[width=\linewidth,height=30.0 em]{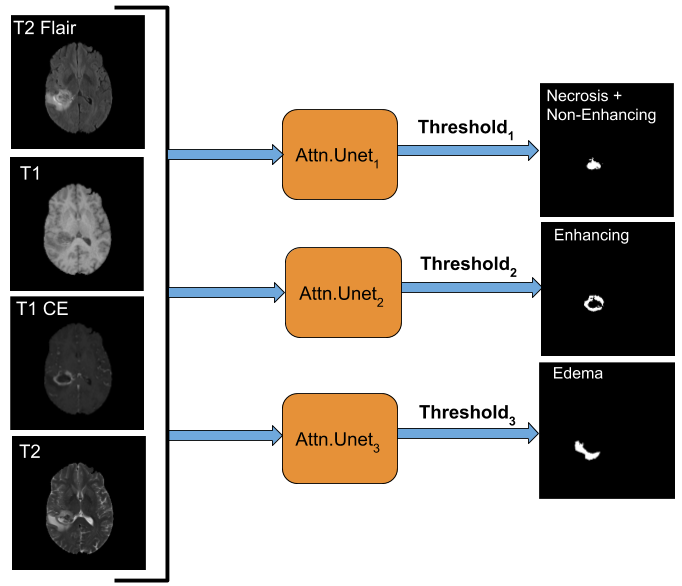}
\caption{\label{Figure-2}{Diagram of the proposed Multi-Threshold Attention U-Net (MTAU) model utilized in this work. Each of the attention U-Nets are independent of each other and their optimally chosen thresholds are numbered as 1,2 and 3 respectively.
}}
\end{figure}

We started by selecting an architecture for 2D Attention U-Net model \cite{oktay2018attention} followed by training this architecture for three different tasks (or regions) and finally ended up with three different models, one for each region viz. NCR+NET, ET and edema. We can also use the same architecture in multi-path format by stacking and training three models in parallel, if there are no constraints on GPU memory. The model architecture is shown in Fig. \ref{Figure-1} with the various connections along with actual parameters involved in each layer. This architecture was replicated and trained for each of the three regions as shown in Fig. \ref{Figure-2}. All three attention U-Nets are independent of each other, hence we can chose a separate threshold for each region. The purpose of threshold in each model is to binarize the output values to 0 and 1. Training separate model for each region gives the flexibility of choosing a separate threshold based on Area Under Curve (AUC) of the respective outputs and thus the model is termed as `Multi-Threshold Attention U-Net (MTAU)'.

\subsection{Dataset}
\begin{figure}[!h]
\centering\includegraphics[width=\linewidth,height=35.0 em]{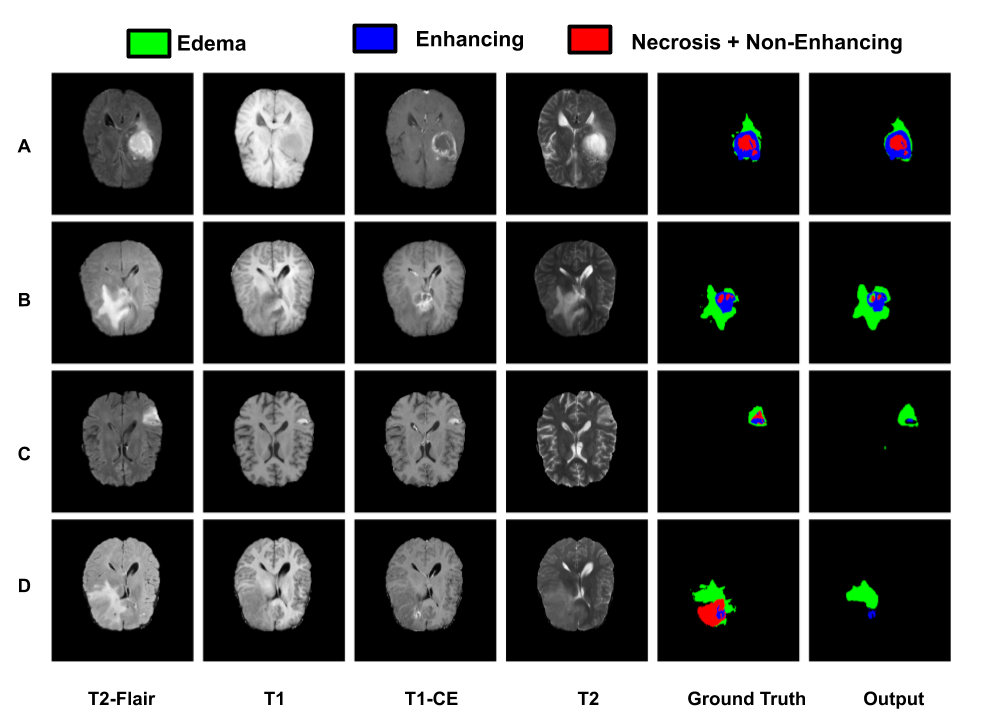}
\caption{\label{Figure-3}{ The segmentation results obtained utilizing the proposed model for the best as well as below average results for the training dataset. Every row represents a different patient data. The columns represent the T2-FLAIR, T1, T1-CE, T2, Ground truth and the Output. The segmentation labels are: Green for Edema, Blue for
Enhancing Tumor and Red for (Necrosis + Non-Enhancing). (A) and (B) represents the best segmentation outputs while (C) and (D) represents the below average segmentation obtained using the proposed model.
}}
\end{figure}

Our proposed technique was trained on BraTS 2020 training dataset \cite{bakas2017advancing,bakas2018identifying}. The training dataset consists of 369 scans while the validation dataset consists of 125 MRI scans of glioblastoma (GBM/HGG) and lower grade glioma (LGG). Also, the dataset has been updated from last year with more routine clinically-acquired 3T multimodal MRI scans and accompanying ground truth labels by expert board-certified neuroradiologists. The various sub -regions considered for the segmentation evaluation are: 1) the ``enhancing tumor" (ET), 2) the ``tumor core" (TC), and 3) the ``whole tumor" (WT). The ET area is shown in hyper-intensity in T1Gd as compared to T1, and also as compared to ``healthy" white matter in T1Gd. Here, TC denotes the bulk of the tumor entailing the ET, necrotic (fluid-filled) and the non-enhancing (solid) parts in the tumor. The appearance of the necrotic (NCR) and the non-enhancing tumor core is typically hypo-intense in T1-Gd as compared to T1. The WT describes the full extent of the pathology by entailing the TC and the peritumoral edema (ED), which is typically shown by hyper-intense signal in FLAIR. The provided segmentation for the dataset have values of 1 for NCR and NET, 2 for ED, 4 for ET, and 0 representing everything else \cite{bakas2017advancing,bakas2018identifying}. An example showing the various regions of interest with the actual segmentation region can be seen in Fig. \ref{Figure-3}.

\begin{figure}[!h]
\centering\includegraphics[width=\linewidth,height=35.0 em]{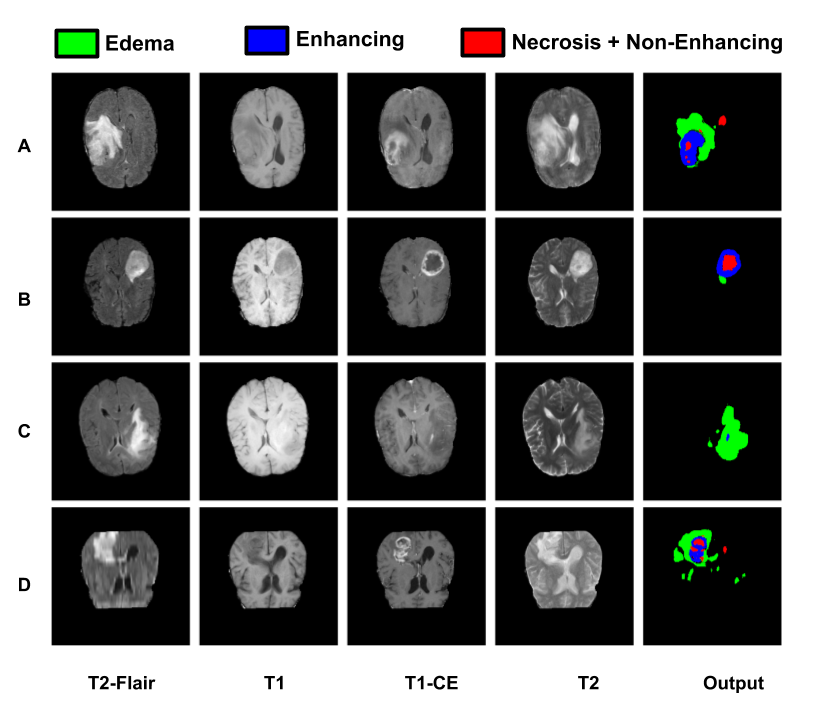}
\caption{\label{Figure-4}{ The segmentation results obtained utilizing the proposed model for the best as well as below average results for the validation dataset. Every row represents a different patient data. The columns represent the T2-FLAIR, T1, T1-CE, T2, Ground truth and the Output. The segmentation labels are: Green for Edema, Blue for
Enhancing Tumor and Red for (Necrosis + Non-Enhancing). (A) and (B) represents the best segmentation outputs while (C) and (D) represents the below average segmentation obtained using the proposed model.
}}
\end{figure}

\subsection{Preprocessing}
The data is first processed to reduce in-homogeneity by scaling each scan volume matrix (of a particular modality) with maximum value in that matrix. Since our approach is 2D segmentation, we create dataset by extracting slices from the volumes. Each volume matrix is sliced into 2D gray scale images. A total of 57,195 (= 369 vols x 155 slices per vol) sets of slices were extracted from the training volumes. Each 2D set comprised of 4 input slices and 3 segmentation maps. The gray scale slices having only zeros in inputs and segmentation maps were removed from the training set. We did this to ensure that results during training do not give a false impression of accuracy while actually model is mapping zeros to zeros. After removal of such slices we were left with 50,899 sets that we split into training, validation, and testing set comprising 40000, 5000,and 5899 sets respectively. The slices of training, validation and test set were taken from different volumes to ensure that model does not over-fit on the given dataset. It is pertinent to mention again that the training, validation and testing set created here were from the 369 training volumes for which ground truth was available. These sets were used for training, validating and testing at our level internally. Apart from this, there is a separate validation set of 125 volumes given by challenge organizers for which ground truth was not provided. The segmentation map on these validation volumes were uploaded on portal to get the performances. The results shown in Table-\ref{table-1} and Table-\ref{table-2} correspond to 369 training volumes and 125 validation volumes as referred by the challenge organizers.

\subsection{Model Architecture}
The model architecture utilized in this work is shown in Fig. \ref{Figure-1}. It consists of the U-Net architecture where the attention gates are used to give more focus to the area of segmentation. The total number of parameters in the network are 3,125,013 consisting of 3,121,141 trainable parameters and 3,872 non-trainable parameters. The learning rate was set to be 1e-5 while the number of epochs were 50, adam was used as an optimizer \cite{kingma2014adam} and the binary cross entropy loss has been used for back-propagation. The model with the best validation set was saved and used for further analysis. The deep learning model was implemented in Keras \cite{chollet2015keras} using Tensorflow \cite{abadi2016tensorflow} as backend. \\
As discussed earlier, instead of using a single model we utilized three different models for training as well as testing of the dataset. We made three models with same architecture for each of the segmentation masks (1 for NCR and NET, 2 for ED, 4 for ET) provided in the dataset. Each of the model is trained independently and then the three models are saved for further analysis. These three models are then used for inference and the results are combined to get the final output of the three segmentation maps. The resulting outputs are further combined into a 3D segmentation map of the output volume. The proposed model has lesser complexity, requires lesser memory as compared to a 3D model and can be trained much faster if the models are trained in parallel.

\section{ {Figures of Merit}}
 {The reliability of the proposed model or method should be checked by means of parameters that can quantify the accuracy and validity of the test results and the proposed model. Following are commonly used metrics in computer vision utilised in this work.}

\subsection{ {Dice Coefficient (DSC)}}
 {This parameter is used to calculate similarity between two sample sets. It is also known as Sørensen–Dice index or Dice Similarity Coefficient and calculated as \cite{carass2020evaluating}:\newline
\[DSC =\frac{2*|X\cap Y|}{|X|+|Y|}\]
where X and Y are the two samples/sample sets.\abs{X} and \abs{Y} represent the cardinalities of set X and Y.}

\subsection{ {Sensitivity (SN)}}
 {This value gives the True positive rates reported by a test. It tells the number of times the test gave a positive result when the sample/person indeed had a condition \cite{altman1994diagnostic}. It is calculated as:\newline
\[SN =\frac{TP}{TP+FN}\]
where TP are the number of true positives and FN are the number of false negatives.}
\subsection{ {Specificity (SP)}}
 {This value gives the True negative rates reported by a test. It tells the proportion of samples reported negative by the test which are indeed negative \cite{altman1994diagnostic}. Specificity is calculated as:\newline
\[SP =\frac{TN}{TN+FN}\]
where TN are the number of true negatives and FN are the number of false negatives.}
\subsection{ {Hausdorff Distance (h)}}
 {This value is a measure of distance between two sample sets. It is the maximum distance of a set from the nearest point in the other set \cite{huttenlocher1993comparing}. Hausdorff Distance is calculated as:\newline
\[h(X,Y)=max_{x\epsilon X} min_{y\epsilon Y}\|x-y\|\]
where X=\{$x_1,x_2,.....,x_n$\} and Y=\{ $y_1,y_2,.......y_n$\} are sample sets.}

\begin{table}[h]
\caption{Performance of proposed method on BraTS 2020 training dataset for segmentation}\label{table-1}
\begin{tabular}{|l|l|l|l|l|l|l|l|l|l|l|l|l|}
\hline
{Evaluation Metrics} &  \multicolumn{3}{|c|}{Dice}  & \multicolumn{3}{|c|}{Sensitivity}   & \multicolumn{3}{|c|}{Specificity} & \multicolumn{3}{|c|}{Hausdorff}\\
 &  ET & WT & TC   &  ET & WT & TC  &  ET & WT & TC   &  ET & WT & TC \\
\hline
Mean &	0.59 &	0.72 &	0.61 &	0.52 &	0.72 &	0.63 &	0.99 &	0.99 &	0.99 &	38.87 &	20.81 &	24.22 \\
Std. Dev. &	0.30 &	0.19 &	0.25 &	0.31 &	0.24	& 0.27 &	0.00 &	0.00 &	0.00 &	100.75 &	20.02 &	28.35\\
Median &	0.70 &	0.78 &	0.67 &	0.57 &	0.80 &	0.71 &	0.99 &	0.99 &	0.99 &	4.33 &	11.70 &	14.09\\
25 Quantile &	0.41 &	0.66 &	0.46 &	0.28 &	0.59 &	0.47 &	0.99 &	0.99 &	0.99 &	2.23 &	7.48 &	8.17\\
75 Quantile &	0.83 &	0.86  &	0.82 &	0.79 &	0.93 &	0.85 &	0.99 &	0.99 &	0.99 &	12.36 &	29.51 &	31.05\\
\hline
\end{tabular}
\end{table}

\begin{table}[h]
\caption{Performance of proposed method on BraTS 2020 validation dataset for segmentation}\label{table-2}
\begin{tabular}{|l|l|l|l|l|l|l|l|l|l|l|l|l|}
\hline
 Evaluation Metrics &  \multicolumn{3}{|c|}{Dice} & \multicolumn{3}{|c|}{Sensitivity}   & \multicolumn{3}{|c|}{Specificity} & \multicolumn{3}{|c|}{Hausdorff} \\
 &  ET & WT & TC   &  ET & WT & TC  &  ET & WT & TC   &  ET & WT & TC \\
\hline
Mean &	0.57 &	0.73 &	0.61 &	0.52 &	0.77	 & 0.62 &	0.99 &	0.99 &	0.99 &	47.22 &	24.03 &	31.53\\
Std. Dev.	& 0.33 &	0.17 &	0.26 &	0.34 &	0.23	& 0.27	& 0.00 &	0.00 &	0.00 &	108.70 &	22.81 &	41.07\\
Median	& 0.71 &	0.78	& 0.69	& 0.60	& 0.86	& 0.70	& 0.99	& 0.99 &	0.99 &	4.12 &	13.74	& 15.84\\
25 Quantile	& 0.30 &	0.67 &	0.38 &	0.18	& 0.62 &	0.47 &	0.99	& 0.99 &	0.99 &	2.23 &	7.00 &	9.69\\
75 Quantile	& 0.84 &	0.84 &	0.82	& 0.81 &	0.95 &	0.83 &	0.99	& 0.99	& 0.99 &	15.66 &	35.76 &	46.79\\
\hline
\end{tabular}
\end{table}

\begin{table}[h]
\caption{ {Performance of proposed method on BraTS 2020 testing dataset for segmentation}}\label{table-3}
\begin{tabular}{|l|l|l|l|l|l|l|l|l|l|l|l|l|}
\hline
 Evaluation Metrics &  \multicolumn{3}{|c|}{Dice} & \multicolumn{3}{|c|}{Sensitivity}   & \multicolumn{3}{|c|}{Specificity} & \multicolumn{3}{|c|}{Hausdorff} \\
 &  ET & WT & TC   &  ET & WT & TC  &  ET & WT & TC   &  ET & WT & TC \\
\hline
Mean &	0.59 &	0.72&	0.57&	0.57&	0.76&	0.60&	0.99&	0.99&	0.99&	38.53&	22.73&	41.82\\
Std. Dev.	& 0.29&	0.18&	0.27&	0.30&	0.20&	0.29&	0.00&	0.00&	0.00&	102.68&	23.44&	78.27\\
Median	& 0.70&	0.79&	0.63&	0.57&	0.81&	0.71&	0.99&	0.99&	0.99&	4.30&	10.00&	16.99\\
25 Quantile	& 0.46&	0.66&	0.47&	0.31&	0.68&	0.40&	0.99&	0.99&	0.99&	2.23&	5.93&	8.12\\
75 Quantile	& 0.82&	0.85&	0.79&	0.80&	0.93&	0.84&	0.99&	0.99&	0.99&	10.28&	33.56&	45.40\\
\hline
\end{tabular}
\end{table}
\section{Results and Discussions}
The performance of the proposed model was tested on the training, validation as well as testing dataset. The training dataset consists of 369 scans while the validation dataset consists of 125 MRI scans of glioblastoma (GBM/HGG) and lower grade glioma(LGG). All computations were carried out on a Linux workstation with Intel Xeon Silver 4110 CPU with 2.10 GHz clock speed, having 128 GB RAM and a TITAN RTX GPU with 24 GB memory. The results are obtained using the proposed model for the training, validation as well as testing dataset. Fig. \ref{Figure-3} shows the segmentation results for training dataset utilized for training the model. Fig. \ref{Figure-3}(A) and (B) represents the best cases while Fig. \ref{Figure-3}(C) and (D) represents the below average results obtained on the training dataset obtained using the proposed segmentation model. The results obtained for the complete training dataset are provided in Table-\ref{table-1}. The dice value obtained for the enhancing tumor, whole tumor and tumor core were found to be 0.59, 0.72, and 0.61 respectively. The values of Std. Dev., Median, 25 Quantile, and 75 Quantile are also provided in Table-\ref{table-1}. The values of sensitivity, specificity, and Hausdorff distance are also calculated and are shown in Table-\ref{table-1}.

Fig. \ref{Figure-4} shows the segmentation results for validation dataset utilized for validation the model. Fig. \ref{Figure-4}(A) and (B) represents the best results while Fig. \ref{Figure-4}(C) and (D) represents the below average results obtained on the validation dataset using the proposed segmentation model. The results obtained for the complete validation dataset are provided in Table-\ref{table-2}. The dice value obtained for the enhancing tumor, whole tumor and tumor core were found to be 0.57, 0.73, and 0.61 respectively. The values of Std. Dev., Median, 25 Quantile, and 75 Quantile are also provided in Table-\ref{table-2}. The values of sensitivity, specificity, and Hausdorff distance are also calculated and are shown in Table-\ref{table-2}. 

 {The results obtained for the complete testing dataset are provided in Table-\ref{table-3}. The dice value obtained for the enhancing tumor, whole tumor and tumor core were found to be 0.59, 0.72, and 0.57 respectively. The values of Std. Dev., Median, 25 Quantile, and 75 Quantile are also provided in Table-\ref{table-3}. The values of sensitivity, specificity, and Hausdorff distance are also calculated and are shown in Table-\ref{table-3}. The training, validation, as well as testing dataset outputs are obtained utilizing the online portal for submission of the results.}

\section{Conclusions}

A multi-threshold model was developed based on attention U-Net for identification of various regions of the tumor in MRI scans. The proposed model offers the advantages of reduced computational complexity, less memory requirements as well as less training time if trained in parallel. The proposed model achieved mean Dice Coefficient of 0.59, 0.72, and 0.61 for enhancing tumor, whole tumor and tumor core respectively on the training dataset.  {The same model gave mean Dice Coefficient of 0.57, 0.73, and 0.61 on the validation dataset and 0.59, 0.72, and 0.57 respectively on the testing dataset.}

\bibliographystyle{splncs04}
\bibliography{my_bib}
\end{document}